\documentclass[runningheads]{llncs}
\usepackage[T1]{fontenc}
\usepackage{graphicx,verbatim}
\usepackage{amsmath}
\usepackage{amssymb}
\usepackage{booktabs}    % for \toprule \midrule \bottomrule
\usepackage{multirow}    % for \multirow
\usepackage[table]{xcolor} % for \rowcolor
\usepackage{float}
\begin{document}

\vbadness=2000000000
\hfuzz=100pt

\setlength{\abovedisplayskip}{0pt}
\setlength{\belowdisplayskip}{0pt}
\setlength{\floatsep}{1pt plus 1.0pt minus 1.0pt}
\setlength{\intextsep}{1pt plus 1.0pt minus 1.0pt}
\setlength{\textfloatsep}{1pt plus 1.0pt minus 1.0pt}
\setlength{\parskip}{1pt}
\setlength{\abovedisplayshortskip}{0pt}
\setlength{\belowdisplayshortskip}{0pt}

%
% \title{Your MICCAI Paper Title}
\title{Test-time Adaptation of Pelvic Bone Segmentation Models via Dynamic Reliability-Guided}
\titlerunning{TTA of PBS models via Dynamic Reliability-Guided}
%\begin{comment}  Removed for anonymized MICCAI submission

\author{Ling Ren\inst{1}\and
Chao Deng\inst{1}\thanks{Corresponding author. \email{dengchao\_neu@126.com}} \and
Ziming Wang\inst{1} \and
Yuecong Xu\inst{2} \and
Kai Zheng\inst{3}}
% index{Ren, Ling}
% index{Deng, Chao}
% index{Wang, Ziming}
% index{Xu, Yuecong}
% index{Zheng, Kai}
\authorrunning{L. Ren et al.}

\institute{
College of Automation, Nanjing University of Posts and Telecommunications,
Nanjing 210003, China\\
\and
Department of Electrical and Computer Engineering, National University of Singapore, Singapore 117583
\and
State Key Laboratory Cultivation Base of Research, Prevention and Treatment for Oral Diseases, the Affiliated Stomatological Hospital of Nanjing Medical University, Nanjing 210029, China
}

%\end{comment}

%\author{Anonymized Authors}  %% Added for anonymized MICCAI submission

%\authorrunning{Anonymized Author et al.}
%\institute{Anonymized Affiliations}
% \institute{Paper ID \ 1573}
%  \\    \email{email@anonymized.com}}
  
\maketitle              % typeset the header of the contribution
\begin{abstract}
Reliable pelvic bone segmentation (PBS) from CT is essential for robot-assisted pelvic trauma surgery, yet deploying a source-trained model to a new hospital suffers from severe performance degradation due to cross-center domain shifts. While test-time adaptation (TTA) enables online model adaptation without accessing source data, existing methods show limited effectiveness for PBS, facing challenges including boundary degradation, anatomical inconsistency under domain shifts, and voxel-level class imbalance. To address these challenges, we propose a novel closed-loop dynamic Reliability-Guided TTA framework (ReGA) for PBS. Specifically, we introduce a pseudo-label reliability criterion termed Segmentation Inference Consistency Evaluation (SICE), which jointly measures region overlap and boundary deviation via dropout-based ensemble predictions. Based on SICE, a trust-weighted refinement module adaptively updates features to mitigate boundary errors in pseudo-labels. Furthermore, a confidence-weighted region-level contrastive learning strategy is proposed to enforce anatomical consistency. Finally, ReGA follows the teacher-student (TS) scheme to alleviate voxel-level class imbalance.
Experiments on three heterogeneous 3D pelvic CT datasets demonstrate that ReGA consistently outperforms state-of-the-art TTA methods, enabling effective adaptation of the source-trained PBS model to unseen clinical domains. The code is available at https://github.com/Ren-ling/ReGA.

% The abstract should briefly summarize the contents of the paper in 150--250 words.  If you are to include a link to your Repository, please make sure it is anonymized for the double-blind review phase.

% Authors must provide keywords and are not allowed to remove this Keyword section.

\end{abstract}
\section{Introduction}
Reliable Pelvic Bone Segmentation (PBS) from CT is crucial for the pre-operative planning and intra-operative navigation of robot-assisted pelvic trauma surgery \cite{liu2025end,liu2025preoperative}.
To avoid costly data annotation and repeated model training, deploying a well-trained PBS model to a new hospital is highly desirable. However, discrepancies in imaging devices and patient population heterogeneity introduce severe domain shifts, which lead to performance degradation of the deployed model in the new hospital. Unsupervised domain adaptation methods mitigate domain shift by knowledge transfer to an unlabeled target domain~\cite{cai2025style,zhang2024mapseg} (i.e., a new hospital), improving deployment performances. However, these methods require full access to source domain data, which may raise privacy concerns~\cite{litrico2023guiding}. 

To cope with the data privacy issue, source-free domain adaptation (SFDA) methods were introduced to avoid accessing any source data during adaptation~\cite{liang2020we}. Yet SFDA methods typically require sufficient target-domain samples collected offline~\cite{yang2022dltta}. In reality, it is usually difficult to obtain such pelvic CT data in a new hospital that aligns with current imaging protocols, as these data arrive in a case-by-case manner. Motivated by test-time adaptation (TTA) that performs continuous online adaptation during the test stage~\cite{wang2020tent}, we introduce it in the adaptation of the deployed PBS model.

Recently, several studies have explored TTA for medical image data~\cite{chen2024each,dong2024medical,wu2023upltta,zhang2025iplc,zhou2026tegda}. Most of them achieve adaptation by modifying normalization layers of source-trained models~\cite{chen2024each,dong2024medical} or teacher-student (TS) schemes based on pseudo labels~\cite{wu2023upltta,zhang2025iplc,zhou2026tegda}. However, these methods exhibit performance degradation when applied in cross-domain PBS due to three major challenges.
First, when deployed in a new hospital, a well-trained PBS model struggles to delineate boundaries between adjacent pelvic bones, which exhibit appearance variations across domains. Consequently, the prevailing TS schemes are confined by false prediction and error accumulation at these boundaries. Second, as observed from the ground truth (GT), enforcing anatomical consistency is crucial for the reliable adaptation of the PBS model. However, existing methods neglect the self-supervised abilities within the target domain to achieve this. Third, PBS faces voxel-level class imbalance, in which the pelvic bones occupy only a small fraction of the volume. The class imbalance causes minor parameter updates during adaptation to be easily dominated by background voxels, impairing segmentation performance.

To cope with these challenges for effective TTA in PBS, we propose a novel closed-loop dynamic \textbf{Re}liability-\textbf{G}uided TT\textbf{A} framework (\textbf{ReGA}). Prior methods exhibit coarse boundary awareness in the target domain, so we propose a novel pseudo-label reliability criterion termed \emph{Segmentation Inference Consistency Evaluation} (SICE). The SICE measures dropout-based ensemble segmentation consistency by considering region overlap and boundary deviation through Dice and Hausdorff distance metrics, further calibrated by confidence scores to improve assessing prediction quality. Based on SICE, a Trust-Weighted Adaptive Feature Refinement (TAFR) module selects high-SICE features from a memory bank, fusing them with the current testing sample to refine pseudo-labels. Furthermore, to enforce anatomical consistency under domain shifts, we propose a Region-Level Contrastive Learning (RCL) module that contrasts trust-aware regional centroids. To this end, ReGA follows the TS scheme, which is commonly used in class imbalance segmentation and provides target-specific supervision.

In summary, our contributions are threefold. First, to the best of our knowledge, we propose the first TTA framework for PBS, named ReGA, which enables effective adaptation to unseen clinical domains.
Second, we propose a closed-loop reliability-guided refinement scheme to address boundary degradation during online adaptation. In addition, a confidence-weighted region-level contrastive loss is introduced to enforce anatomical feature consistency across pelvic regions. Finally, extensive experiments on three 3D pelvic CT datasets demonstrate that ReGA outperforms state-of-the-art TTA methods.

\begin{figure*}[t] 
 \centering 
\includegraphics[width=.85\textwidth]{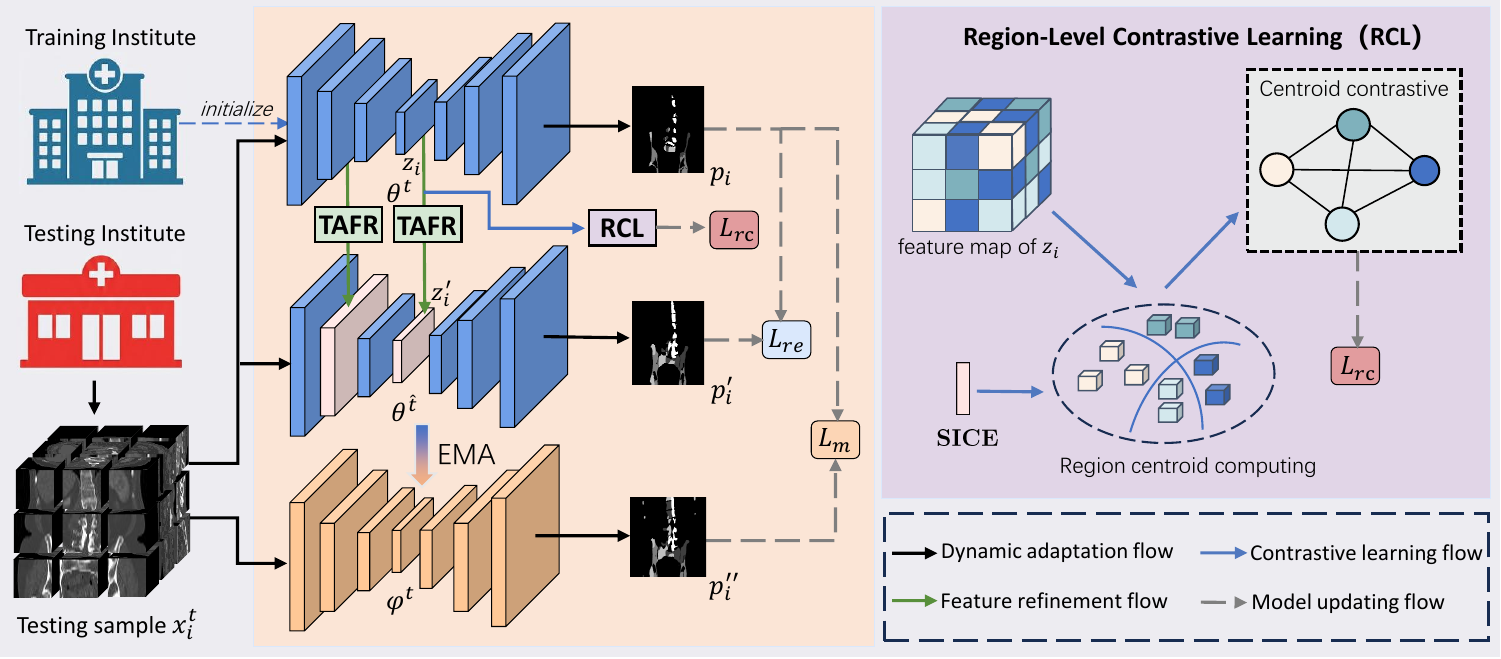}
\caption{Overall process of the proposed ReGA. 
During adaptation, the features of a testing sample $x_i^{t}$ extracted by the encoder of the PBS model $\theta^{t}$ (initialized at the training institute) are refined by TAFR (see Fig.~\ref{component}). The refined features then replace those of the model $\theta^{\hat{t}}$ copied from $\theta^{t}$ to generate refined predictions. With the SICE score from CSCS (see Fig.~\ref{component}) and the feature map of $z_i$, region centroids are computed for region-level contrastive learning. Finally, $\theta^{t}$ is optimized by minimizing the refined pseudo-label loss $L_{re}$, mean teacher loss $L_m$ and contrastive loss $L_{rc}$.}
\label{overall} 
\end{figure*}

\begin{figure*}
\centering 
\includegraphics[width=.85\textwidth]{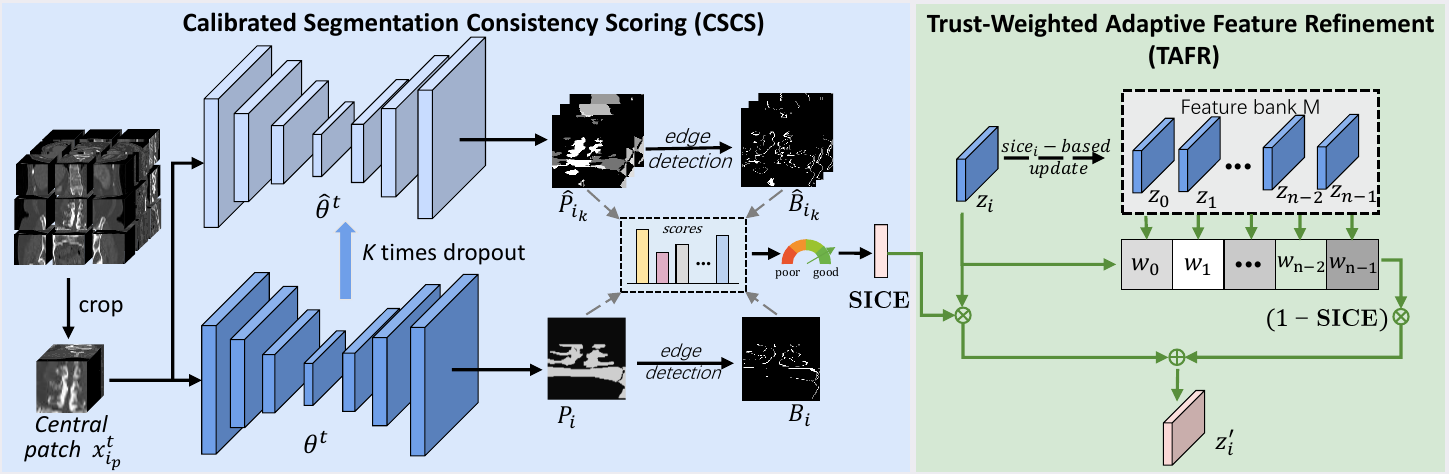}
\caption{Overview of the CSCS and TAFR module. In CSCS, the test sample $x_i^{t}$ is centrally cropped to obtain $x_{i_p}^{t}$, which is passed through $\theta^t$ with $K$-times dropout for multiple inferences. These inferences, along with the edges extracted using the Canny operator, are combined to compute SICE. In TAFR, latent features of high SICE are dynamically accumulated into a feature bank $M$. The current feature is then adaptively refined through weighted fusion with those from the bank, guided by SICE and cosine similarity, to achieve adaptation. }
\label{component} 
\end{figure*}

\section{Methodology}

Given a PBS model $\theta^{t}$ trained on a source domain dataset $\mathcal{D}^s =\{(x_i^s, y_i^s)\}_{i=1}^{N^s}$, where $x_i^s$ is a 3D tensor and $y_i^s$ is a densely labeled 3D segmentation mask, with $N^s$ representing the number of samples in the source domain. Let the unlabeled target domain dataset for testing be denoted as $\mathcal{D}^t=\{x_i^t\}_{i=1}^{N^t}$ and $N^t$ represents the number of samples in the target domain, which has a distribution shift from $\mathcal{D}^s$. 
The goal of this paper is to adapt $\theta^t$ in an online manner within each batch $B^j = \{x_i^t\}_{i=1}^B$ from $\mathcal{D}^t$, where $j \in {1, \dots, N^t / B}$ and $B$ denotes the batch size. To achieve this, we propose ReGA for the adaptation of the PBS model on $\mathcal{D}^t$, as illustrated in Fig.~\ref{overall}.

\subsection{Calibrated Segmentation Consistency Scoring}
The performance of PBS models is commonly evaluated using the Dice coefficient and the Hausdorff distance. In real-world inference, where GT for test samples are unavailable, automated performance evaluation becomes crucial. Although Monte Carlo (MC) dropout~\cite{gal2016dropout} offers uncertainty estimates through multiple forward passes, these uncertainty maps do not directly reflect the segmentation accuracy, particularly of boundary delineation.

Motivated by prior studies~\cite{lee2024aetta,zhou2026tegda} that demonstrate a correlation between prediction variance under dropout and testing errors, we propose Calibrated Segmentation Consistency Scoring (CSCS) to calculate SICE to evaluate the reliability of pseudo-labels generated by the target model during adaptation. SICE introduces a boundary-oriented indicator to provide a more precise assessment of segmentation quality.

Specifically, let $\hat{\theta}_k^{t}$ denote the target model $\theta^{t}$ parameters with the $k$-th MC dropout.
As shown in Fig.~\ref{component}, given a test sample $x_i^t$, the central patch $x_{i_p}^t$ is fed into $\theta^t$ and its $k$-th dropout version $\hat{\theta}_k^{t}$, yielding predictions $P_i = f(\theta^t, x_{i_p}^t)$ and $\hat{P}_{i_k} = f(\hat{\theta}_k^{t}, x_{i_p}^t)$, respectively. To explicitly characterize boundary discrepancies, edge maps are extracted from $P_i$ and $\hat{P}{i_k}$ using the Canny operator~\cite{yang2025boundary}, producing $B_i = \mathrm{Canny}(P_i)$ and $\hat{B}_{i_k} = \mathrm{Canny}(\hat{P}{i_k})$.
The proposed SICE score is formulated as a weighted combination of the volumetric segmentation consistency ($S_v$) and the boundary discrepancy ($S_b$). The volumetric consistency metric is defined as
\begin{equation}
\mathcal{S}(x,\hat{x}) =
\frac{1}{K \cdot L
}
\sum_{k=1}^{K}
\sum_{l=0}^{L-1}
\frac{
2 \sum_{v \in  \mathrm{\Omega}} x_{v}^l \cdot \hat{x}^{l}_{k,v}}{\sum_{v \in \mathrm{\Omega}} x_{v}^l
+\sum_{v \in \mathrm{\Omega}} \hat{x}^{l}_{k,v}
},\end{equation}
where $\mathrm{\Omega}$ represents the set of voxel indices, with $v = (h, w, d) \in  \mathrm{\Omega}$ denoting the index of each voxel, $K$ is the number of dropout forward passes and $L$ is the number of classes. $x_v^{l}$ and $\hat{x}_{k,v}^l$ denote the predicted probability and those obtained by the $k$-th dropout version for the $l$-th class of the $v$-th voxel, respectively. 

By instantiating $\mathcal{S}(\cdot,\cdot)$ with the segmentation outputs, the volumetric segmentation consistency is obtained as
\begin{equation}
S_v = \mathcal{S}(P_i, \hat{P}_{i_k}).
\end{equation}
Subsequently, the boundary consistency measure \( S_b \) is computed as
\begin{equation}
S_b = \max \left( P_{\gamma}(B_i, \hat{B}_{i_k}),\, P_{\gamma}(\hat{B}_{i_k},B_i) \right),
\end{equation}
where $P_{\gamma}(B_i, \hat{B}_{i_k})$ denotes the $\gamma$ percentile of the minimum distances from all voxels in the boundary voxel sets $B_i$ to set $\hat{B}_{i_k}$, and $P_{\gamma}(\hat{B}_{i_k}, B_i)$ denotes the calculation in the revise direction. Finally, the reliability score $R_{i}$ for the current testing sample $x_i^t$ is defined as
\begin{equation}
\label{eq:reliability}
R_{i} = \lambda_1 S_v + \lambda_2 \exp\left(-S_b / \alpha\right),
\end{equation} where $\lambda_1$ and $\lambda_2$ weight region- and boundary-level consistency, respectively, and $\alpha$ modulates the sensitivity to boundary deviations.

The dropout-based inference often exhibits high consensus in the interior regions of the target domain predictions, whereas discrepancies are primarily concentrated at the boundaries~\cite{wu2024fpl+}. Consequently, $S_v$ tends to overestimate the true segmentation performance. Thus, we introduce a calibrating coefficient $\lambda \in (0, 1)$ to calibrate the estimation based on the overall confidence as follows
\begin{equation} \lambda = 1 - \frac{1}{|\mathrm{\Omega}| \log C} \sum_{v \in \mathrm{\Omega}} \left( - \sum_{l=0}^{L} \bar{P}_{v,l} \log \bar{P}_{v,l} \right), \label{eq:calibration_factor} \end{equation}
where $\bar{P} = \frac{1}{K} \sum_{k=1}^{K}\hat{P}_{i_k}$ denotes the ensemble prediction map and $\log C$ is the maximum entropy value used for normalization. The proposed SICE of the current testing sample $x_i^t$ is formulated as ${SICE}_{i} = \lambda \cdot R_{i}$, which is used to select high-confidence samples and guide the subsequent adaptation process.

\subsection{Trust-Weighted Adaptive Feature Refinement}
As shown by the previous method~\cite{zheng2024dual}, explicit feature alignment from well-predicted samples helps minimize the domain gap for other testing samples. Thus, we propose a trust-weighted refinement module with a dynamic feature bank that stores high-quality target domain representations and applies trust-weighted fusion for feature correction.

As described in Fig.~\ref{component}, the dynamic feature bank $M$ comprises $\{Z_i\}_{i=0}^{n-1}$, where $Z_i=f(\theta^{t}_e,x_i^t)$ and $M$ are updated using a first-in-first-out principle to ensure adaptability to the distribution of the incoming test sample. Specifically, when $\mathrm{SICE}_{i}$ exceeds the $\tau$ percentile of $\mathrm{SICE}$ values of all previous samples before time step $t$, $Z_i$ is incorporated into the feature bank $M$. 

For a new testing sample $x_i^t$, the cosine similarity between $Z_i$ and the $j_{th}$ feature $Z_j$ in $M$ is computed as $sim(Z_i,Z_j)$. Then $Z_i$ is updated as a trust-weighted combination of $Z_i$ and the feature $Z_f$, where $Z_f = \sum_{j=0}^{n-1} W_{j} \cdot Z_j$ represents the reference feature from the dynamic bank based on the cosine similarity and $W_{j} = sim(Z_i, Z_j) / \sum_j sim(Z_i, Z_j)$. To encourage the retention of original features for well-predicted samples, while replacing those with poor predictions, the refined feature $Z_i'$ is defined as
\begin{equation}
Z_i' =  w(SICE_i) \cdot Z_i + (1 - w(SICE_i)) \cdot Z_f,
\end{equation}
where $w(\cdot)$ represents the normalization operation. The refined feature $Z_i'$ is then sent to the student model $\theta^{\hat{t}}$ to obtain a refined prediction $p_i'$.

\subsection{Region-Level Contrastive Learning}
In the context of domain adaptation for segmentation models, prior works~\cite{yu2023source,zhang2023satta} have utilized pseudo labels to perform contrastive learning for a more compact target feature distribution. With this intuition, we propose the RCL module to enforce anatomical consistency. Rather than relying on voxel-level representations, the region centroid is utilized to represent the entire region for reducing computational cost. Additionally, we dynamically assign weights to voxels based on prediction confidence and refined with the SICE score.
Formally, the confidence-weighted centroid of class $l$ is computed as
\begin{equation}
\mathbf{c}_l = \frac{\sum_{j \in \mathrm{\Omega}} \mathbf{f}_j(Z_i) \cdot \text{I}(\hat{y}_j = l) \cdot (1 - E_i) \cdot SICE_i(l) }{\sum_{j \in \Omega} \text{I}(\hat{y}_j = l)},
\end{equation}
where $\mathbf{f}_j$ is the feature map of $Z_i$ at voxel $j$, $\text{I}(\cdot)$ is the indicator function, $\hat{y}_j$ is the prediction label from the target model $\theta^t$ at voxel $j$, $E_i$ is the entropy-based uncertainty score, and $SICE_i(l)$ denotes the SICE sore for class $l$.

Then, the learning objective for anatomical consistency is formulated as
\begin{equation}
\mathcal{L}_{rc} = - \frac{1}L \sum_{l=0}^{L-1} \log \frac{\exp(\text{sim}(\mathbf{c}_l, \mathbf{c}_l))}{\sum_{m=0, m\neq l}^{L-1} \exp(\text{sim}(\mathbf{c}_l, \mathbf{c}_m))},
\end{equation}
where $\text{sim}(\mathbf{u}, \mathbf{v}) = (\mathbf{u}^\top \mathbf{v}) / (\|\mathbf{u}\| \|\mathbf{v}\| \cdot T)$ denotes the cosine similarity scaled by a temperature parameter $T$. 

\subsection{Self-Adaptive Model Updating}
Although TAFR contributes to producing generally reliable pseudo labels, the gradients can vary significantly across different testing batches, which causes instability during adaptation. In line with established protocol in TTA~\cite{wang2020tent,wang2022continual}, we adopt a mean teacher model to improve stability. At the time step $t = 0$, the mean teacher model  $\phi^t$ is initialized to be the same as the source pre-trained model $\theta^{t}$, i.e., $\phi^0=\theta^{0}$.
Since the traditional mean teacher updated with a constant Exponential Moving Average (EMA) rate fails to effectively handle the dynamic changes in data quality during TTA, we propose a SICE-aware updating rule with an adaptive EMA rate as follows
\begin{equation}
\phi^{t+1} = (1 - SICE_i)\cdot\phi^t+ SICE_i \cdot \theta^{t+1},
\end{equation}
where $\theta^{t+1}$ are the updated student model at the current adaptation step $t$. 

To handle potential noise in the refined prediction $p_i'$, the loss is weighted by $\mathrm{SICE}_i$ to suppress the contribution of poorly adapted samples. The total training loss for our proposed ReGA is defined as
\begin{equation}
    L_{ReGA} = \frac{1}{B} \sum_{i=1}^B \left( L_{m}(p_i'', p_i) + SICE_i \cdot L_{re}(p_i', p_i) + \beta L_{rc} \right),
\end{equation}
where $p_i''$ denotes teacher predictions, $L_m$ and $L_{re}$ correspond to the mean teacher and refined pseudo-label losses implemented with Dice and cross-entropy, and $\beta$ is a tradeoff hyperparameter. For each batch, a back-propagation step is performed using $L_{ReGA}$, followed by a forward pass with the updated student model $\theta^{t+1}$ to generate the final segmentation results.

\begin{table*}
\centering
\caption{Comparison of different TTA methods for cross-domain PBS.}
\label{com}
\footnotesize
\setlength{\tabcolsep}{1.4pt}
\renewcommand{\arraystretch}{0.9}

% \resizebox{\textwidth}{!}{%
\begin{tabular}{l c c c c c c}
\toprule
\multirow{2}{*}{Methods}
& \multicolumn{3}{c}{CLINIC$\rightarrow$KITS19}
& \multicolumn{3}{c}{CLINIC$\rightarrow$MSD$_{\text{T10}}$} \\
\cmidrule(lr){2-4}\cmidrule(lr){5-7}
& Dice $\uparrow$ & HD95 $\downarrow$ & ASD $\downarrow$
& Dice $\uparrow$ & HD95 $\downarrow$ & ASD $\downarrow$ \\
\midrule

Source Only
& $19.7{\pm}9.5$ & $112.7{\pm}20.8$ & $67.2{\pm}24.6$
& $13.8{\pm}6.9$ & $149.3{\pm}25.7$ & $56.3{\pm}22.2$ \\

Target Only
& $97.8{\pm}4.6$ & $32.2{\pm}16.5$ & $18.7{\pm}12.1$
& $91.5{\pm}5.1$ & $33.7{\pm}13.5$ & $19.4{\pm}9.2$ \\

TENT~\cite{wang2020tent}
& $24.9{\pm}7.9$ & $190.6{\pm}66.1$ & $62.9{\pm}27.4$
& $25.8{\pm}12.7$ & $171.0{\pm}53.3$ & $67.4{\pm}25.9$ \\

InTEnt~\cite{dong2024medical}
& $23.4{\pm}6.2$ & $210.6{\pm}87.1$ & $65.4{\pm}30.2$
& $25.1{\pm}10.0$ & $171.9{\pm}61.3$ & $69.4{\pm}25.6$ \\

VPTTA~\cite{chen2024each}
& $24.3{\pm}6.1$ & $200.7{\pm}72.6$ & $63.6{\pm}31.1$
& $25.8{\pm}12.6$ & $172.0{\pm}53.6$ & $68.0{\pm}29.8$ \\

CoTTA~\cite{wang2022continual}
& $24.6{\pm}7.6$ & $192.4{\pm}63.0$ & $64.1{\pm}29.0$
& $25.1{\pm}11.6$ & $159.8{\pm}62.0$ & $63.2{\pm}30.4$ \\

TEGDA~\cite{zhou2026tegda}
& $24.3{\pm}10.0$ & $187.7{\pm}67.3$ & $63.4{\pm}27.7$
& $25.8{\pm}10.3$ & $168.0{\pm}63.2$ & $62.1{\pm}29.3$ \\

\midrule
\rowcolor{gray!15}
\textbf{Ours}
& $\textbf{32.1}{\pm}\textbf{7.3}$ & $\textbf{160.4}{\pm}\textbf{42.8}$ & $\textbf{57.9}{\pm}\textbf{29.5}$
& $\textbf{26.4}{\pm}\textbf{12.0}$ & $\textbf{143.9}{\pm}\textbf{70.1}$ & $\textbf{54.2}{\pm}\textbf{35.4}$ \\
\bottomrule
\end{tabular}
% }
\end{table*}

\section{Experiments and Results}
\textbf{Datasets and Implementation Details.}
% \subsubsection{Datasets and Implementation Details.}
We extensively evaluate the proposed ReGA on cross-domain PBS with three sub-datasets of the public dataset CTPelvic1K~\cite{liu2021deep}, including 1) MSD\_T10: 155 cases collected from the 10th sub-dataset of Medical Segmentation Decathlon~\cite{simpson2019large}, 2) KITS19: 44 pelvic CT scans comes from the Kits19 challenge~\cite{heller2019kits19}, and 3) CLINIC: 103 pelvic CT scans collected from an orthopedic hospital without metal artifacts. The three datasets comprise four segmentation classes: sacrum, left hip, right hip, and lumbar spine.

The source model was trained for 400 epochs using the cascaded 3D UNet from nnUNet~\cite{isensee2021nnu}, and the best validation checkpoint was adopted for adaptation. In ReGA, we set a dropout rate of 0.5, the batch size $B=1$, the dropout number $K = 5$, the feature bank length $M = 10$, $\gamma = 95$ for calculating $S_b$, $\tau = 70$ for sample filtering in TAFR, and the tradeoff hyperparameter $\beta = 0.01$. The evaluation metrics are volume-level Dice coefficient, 95th percentile of Hausdorff Distance (HD95), and Average Symmetric Surface Distance (ASD). All models are trained on a single NVIDIA GeForce RTX 4090 24 GB GPU. 

\noindent\textbf{Comparison with State-of-the-art TTA Methods.}
% \subsubsection{Comparison with State-of-the-art TTA Methods.} 
We selected five state-of-the-art methods for detailed comparison, including normalization-based methods TENT~\cite{wang2020tent}, InTEnt~\cite{dong2024medical}, and VPTTA~\cite{chen2024each}; as well as TS schemes CoTTA~\cite{wang2022continual} and TEGDA~\cite{zhou2026tegda}. We also report results for the source-only model, obtained by applying the source pre-trained model directly to the target data. Table~\ref{com} presents the quantitative results for all comparative methods across different source and target domains. In the CLINIC$\rightarrow$KITS19 setting, the source model achieves a Dice coefficient of only 19.7\%. In contrast, state-of-the-art TTA methods range from 23.4\% to 24.9\%, while our ReGA method achieves the highest mean Dice coefficient of 32.1\%, a significant improvement of 12.4\% over the source model, and outperforms other methods. Moreover, our proposed ReGA achieves the lowest HD95 (160.4) and ASD (57.9) values, indicating improved boundary precision and structural consistency.

\noindent\textbf{Ablation Study.} We conduct an ablation study in Table~\ref{ablation} to evaluate the effectiveness of ReGA. Specifically, w/o $L_{m}$, w/o $L_{re}$, and w/o $L_{rc}$  
correspond to cases in which the mean-teacher loss, the TAFR module, and the RCL module are disabled during adaptation, respectively. Besides, the variant w/o $B$ indicates that $S_b$ of the SICE score is removed, while w/o $E$ denotes that entropy is used in place of SICE. Ablation results across the three datasets consistently show that each module incrementally enhances segmentation performance. From the obvious improvement in HD95 and ASD compared to the variant w/o $B$ and w/o $E$, the proposed SICE can successfully evaluate pseudo-label quality, which contributes to the adaptation process.

\begin{table*}[t]
\centering
\caption{Ablation studies of ReGA for cross-domain PBS.}
\label{ablation}
\footnotesize
\setlength{\tabcolsep}{1.4pt}
\renewcommand{\arraystretch}{0.9}

\begin{tabular}{l c c c c c c}
\toprule
\multirow{2}{*}{Variants}
& \multicolumn{3}{c}{CLINIC$\rightarrow$KITS19}
& \multicolumn{3}{c}{CLINIC$\rightarrow$MSD$_{\text{T10}}$} \\
\cmidrule(lr){2-4}\cmidrule(lr){5-7}
& DSC $\uparrow$ & HD95 $\downarrow$ & ASD $\downarrow$
& DSC $\uparrow$ & HD95 $\downarrow$ & ASD $\downarrow$ \\
\midrule

w/o $L_{m}$
& $31.4{\pm}5.4$ & $221.5{\pm}99.3$ & $62.2{\pm}37.4$
& $25.9{\pm}12.9$ & $163.9{\pm}51.7$ & $63.5{\pm}25.4$ \\

w/o $L_{re}$
& $32.0{\pm}6.1$ & $229.6{\pm}105.8$ & $61.9{\pm}41.2$
& $25.4{\pm}5.7$ & $168.9{\pm}51.7$ & $66.3{\pm}23.2$ \\

w/o $L_{rc}$
& $31.5{\pm}6.1$ & $224.1{\pm}102.5$ & $64.8{\pm}38.8$
& $25.6{\pm}5.4$ & $170.7{\pm}47.9$ & $68.4{\pm}27.3$ \\

w/o $B$
& $31.1{\pm}5.8$ & $225.1{\pm}103.1$ & $65.1{\pm}37.1$
& $25.3{\pm}12.8$ & $166.6{\pm}54.9$ & $66.4{\pm}26.8$ \\

w/o $E$
& $31.4{\pm}6.0$ & $224.4{\pm}102.7$ & $65.0{\pm}38.7$
& $25.1{\pm}12.7$ & $167.6{\pm}54.8$ & $67.9{\pm}27.0$ \\

\midrule
\rowcolor{gray!15}
\textbf{Ours}
& $\textbf{32.1}{\pm}\textbf{7.3}$ & $\textbf{160.4}{\pm}\textbf{42.8}$ & $\textbf{57.9}{\pm}\textbf{29.5}$
& $\textbf{26.4}{\pm}\textbf{12.0}$ & $\textbf{143.9}{\pm}\textbf{70.1}$ & $\textbf{54.2}{\pm}\textbf{35.4}$ \\
\bottomrule
\end{tabular}
\end{table*}

\begin{figure*}[t] 
 \centering 
\includegraphics[width=.85\textwidth]{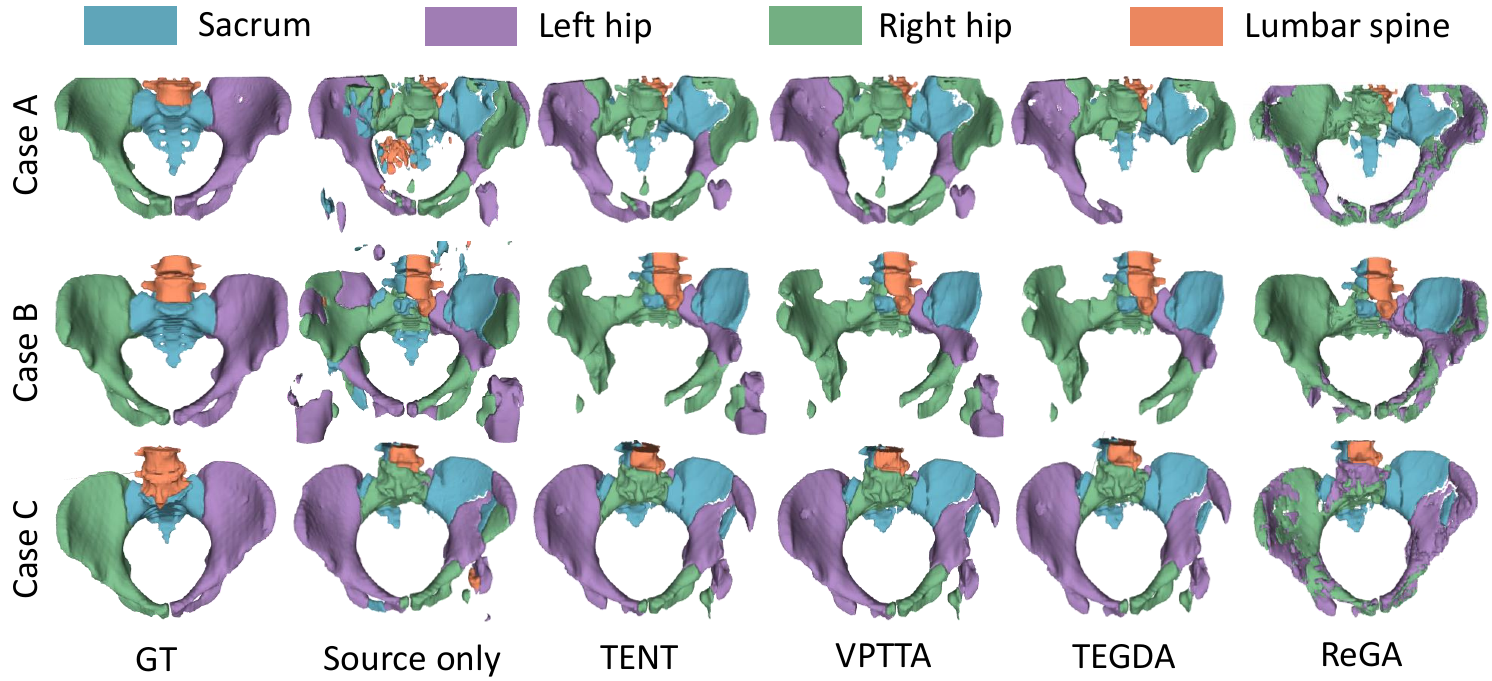}
\caption{Visualization of segmentation results for different TTA methods.}
\label{visualization} 
\end{figure*}
\noindent\textbf{Visualization.}
Fig.~\ref{visualization} presents qualitative comparisons across cases A and C on CLINIC$\rightarrow$KITS19, and case B on CLINIC$\rightarrow$MSD$_{\text{T10}}$.
Existing TTA methods (e.g., TENT, VPTTA, and TEGDA) struggle to maintain structural consistency and boundary accuracy under domain shifts. In contrast, ReGA better preserves anatomical integrity while achieving sharper boundary delineation. Notably, in Cases A and C, competing methods suffer from left–right inversion errors, while ReGA maintains anatomically consistent predictions. In Case B, ReGA demonstrates improved delineation of the right hip bone, better preserving the structural completeness of the pelvis.

\section{Conclusion}
In this paper, we propose ReGA, a dynamic reliability-guided test-time adaptation framework that addresses three key challenges in cross-domain PBS, i.e., boundary degradation, anatomical inconsistency, and voxel-level class imbalance. By combining pseudo-label reliability estimation with adaptive feature refinement and anatomical consistency enforcement, ReGA achieves online adaptation for the PBS model deployment in new clinical settings. Experimental results on three heterogeneous 3D pelvic CT datasets demonstrate that ReGA outperforms current state-of-the-art methods, highlighting its potential for clinical applications. Future work will focus on refining the boundary extraction strategy to
enhance its reliability on more complex anatomical structures.

 %% removed for anonymized MICCAI submission.
    
    % The following acknowledgement and disclaimer sections can be removed for the double-blind review process.  If and when your paper is accepted, reinsert the acknowledgement and the disclaimer clause in your final camera-ready version.
    % IF you opted to include the acknowledgement and disclaimer sections, they will count towards the 8-page limit.

% \clearpage
\bibliographystyle{splncs04}
\bibliography{Paper-1573}

@inproceedings{litrico2023guiding,
  title={Guiding pseudo-labels with uncertainty estimation for source-free unsupervised domain adaptation},
  author={Litrico, M. and Del Bue, A. and Morerio, P.},
  booktitle={CVPR},
  pages={7640--7650},
  year={2023}
}

@article{liu2025preoperative,
  title={Preoperative fracture reduction planning for image-guided pelvic trauma surgery: a comprehensive pipeline with learning},
  author={Liu, Y. and others},
  journal={Medical Image Analysis},
  volume={102},
  number={103506},
  year={2025}
}

@article{liu2025end,
  title={An end-to-end geometry-based pipeline for automatic preoperative surgical planning of pelvic fracture reduction and fixation},
  author={Liu, J. and Li, H. and Zeng, B. and Wang, H. and Kikinis, R. and Joskowicz, L. and Chen, X.},
  journal={IEEE Transactions on Medical Imaging},
  volume={44},
  number={1},
  pages={79--91},
  year={2025}
}

@article{yang2022dltta,
  title={DLTTA: dynamic learning rate for test-time adaptation on cross-domain medical images},
  author={Yang, H. and Chen, C. and Jiang, M. and Liu, Q. and Cao, J. and Heng, P.A. and Dou, Q.},
  journal={IEEE Transactions on Medical Imaging},
  volume={41},
  number={12},
  pages={3575--3586},
  year={2022}
}

@article{wang2020tent,
  title={Tent: fully test-time adaptation by entropy minimization},
  author={Wang, D. and Shelhamer, E. and Liu, S. and Olshausen, B. and Darrell, T.},
  journal={arXiv preprint arXiv:2006.10726},
  year={2020}
}

@inproceedings{dong2024medical,
  title={Medical image segmentation with intent: Integrated entropy weighting for single image test-time adaptation},
  author={Dong, H. and Konz, N. and Gu, H. and Mazurowski, M.A.},
  booktitle={CVPR},
  pages={5046--5055},
  year={2024}
}

@inproceedings{chen2024each,
  title={Each test image deserves a specific prompt: continual test-time adaptation for 2D medical image segmentation},
  author={Chen, Z. and Pan, Y. and Ye, Y. and Lu, M. and Xia, Y.},
  booktitle={CVPR},
  pages={11184--11193},
  year={2024}
}

@inproceedings{wang2022continual,
  title={Continual test-time domain adaptation},
  author={Wang, Q. and Fink, O. and Van Gool, L. and Dai, D.},
  booktitle={CVPR},
  pages={7201--7211},
  year={2022}
}

@article{zhang2025iplc,
  title={IPLC+: SAM-guided iterative pseudo label correction for source-free domain adaptation in medical image segmentation},
  author={Zhang, G. and Qi, X. and Wu, J. and Yan, B. and Wang, G.},
  journal={IEEE Journal of Biomedical and Health Informatics},
  volume={29},
  number={12},
  pages={9060--9072},
  year={2025}
}

@inproceedings{zhou2026tegda,
  title={TEGDA: test-time evaluation-guided dynamic adaptation for medical image segmentation},
  author={Zhou, Y. and Wu, J. and Liao, W. and Zhang, S. and Zhang, S. and Wang, G.},
  booktitle={MICCAI 2025},
  series={LNCS},
  volume={15965},
  pages={628--637},
  year={2026},
  publisher={Springer},
  doi={10.1007/978-3-032-04978-0_60}
}

@inproceedings{wu2023upltta,
  title={UPL-TTA: Uncertainty-Aware Pseudo Label Guided Fully Test Time Adaptation for Fetal Brain Segmentation},
  author={Wu, J. and Gu, R. and Lu, T. and Zhang, S. and Wang, G.},
  booktitle={IPMI 2023},
  series={LNCS},
  volume={13939},
  pages={240--252},
  year={2023},
  publisher={Springer},
  doi={10.1007/978-3-031-34048-2_19}
}

@article{wu2024fpl+,
  title={FPL+: Filtered pseudo label-based unsupervised cross-modality adaptation for 3D medical image segmentation},
  author={Wu, J. and others},
  journal={IEEE Transactions on Medical Imaging},
  volume={43},
  number={9},
  pages={3098--3109},
  year={2024},
  publisher={IEEE}
}

@article{cai2025style,
title={Style mixup enhanced disentanglement learning for unsupervised domain adaptation in medical image segmentation},
author={Cai, Z. and Xin, J. and You, C. and Shi, P. and Dong, S. and Dvornek, N.C. and Zheng, N. and Duncan, J.S.},
journal={Medical Image Analysis},
volume={101},
number={103440},
year={2025},
}

@inproceedings{lee2024aetta,
  title={AETTA: label-free accuracy estimation for test-time adaptation},
  author={Lee, T. and Chottananurak, S. and Gong, T. and Lee, S.J.},
  booktitle={CVPR},
  pages={28643--28652},
  year={2024}
}

@inproceedings{zhang2024mapseg,
  title={MAPSeg: Unified Unsupervised Domain Adaptation for Heterogeneous Medical Image Segmentation Based on 3D Masked Autoencoding and Pseudo-Labeling},
  author={Zhang, X. and others},
  booktitle={CVPR},
  pages={5851--5862},
  year={2024}
}

@inproceedings{gal2016dropout,
  title={Dropout as a bayesian approximation: representing model uncertainty in deep learning},
  author={Gal, Y. and Ghahramani, Z.},
  booktitle={ICML},
  pages={1050--1059},
  year={2016},
  organization={PMLR}
}

@article{yang2025boundary,
  title={Boundary-guided contrastive learning for semi-supervised medical image segmentation},
  author={Yang, Y. and Zhuang, J. and Sun, G. and Wang, R. and Su, J.},
  journal={IEEE Transactions on Medical Imaging},
  volume={44},
  number={7},
  pages={2973--2988},
  year={2025}
}

@article{zheng2024dual,
  title={Dual domain distribution disruption with semantics preservation: Unsupervised domain adaptation for medical image segmentation},
  author={Zheng, B. and others},
  journal={Medical Image Analysis},
  volume={97},
  number={103275},
  year={2024}
}

@inproceedings{zhang2023satta,
  title={SATTA: semantic-aware test-time adaptation for cross-domain medical image segmentation},
  author={Zhang, Y. and Huang, K. and Chen, C. and Chen, Q. and Heng, P.A.},
  booktitle={MICCAI 2023},
  series={LNCS},
  volume={14221},
  pages={160--171},
  year={2023},
  publisher={Springer},
  doi={10.1007/978-3-031-43895-0_14}
}

@inproceedings{yu2023source,
  title={Source-free domain adaptation for medical image segmentation via prototype-anchored feature alignment and contrastive learning},
  author={Yu, Q. and Xi, N. and Yuan, J. and Zhou, Z. and Dang, K. and Ding, X.},
  booktitle={MICCAI 2023},
  series={LNCS},
  volume={14226},
  pages={1--12},
  year={2023},
  publisher={Springer},
  doi={10.1007/978-3-031-43990-2_1}
}

@article{liu2021deep,
  title={Deep learning to segment pelvic bones: large-scale CT datasets and baseline models},
  author={Liu, P. and others},
  journal={IJCARS},
  volume={16},
  number={5},
  pages={749--756},
  year={2021}
}

@article{simpson2019large,
  title={A large annotated medical image dataset for the development and evaluation of segmentation algorithms},
  author={Simpson, A.L. and others},
  journal={arXiv preprint arXiv:1902.09063},
  year={2019}
}

@article{heller2019kits19,
  title={The kits19 challenge data: 300 kidney tumor cases with clinical context, CT semantic segmentations, and surgical outcomes},
  author={Heller, N. and others},
  journal={arXiv preprint arXiv:1904.00445},
  year={2019}
}

@article{isensee2021nnu,
  title={nnU-Net: a self-configuring method for deep learning-based biomedical image segmentation},
  author={Isensee, Fabian and Jaeger, Paul F and Kohl, Simon AA and Petersen, Jens and Maier-Hein, Klaus H},
  journal={Nature methods},
  volume={18},
  number={2},
  pages={203--211},
  year={2021},
}

@inproceedings{liang2020we,
  title={Do we really need to access the source data? source hypothesis transfer for unsupervised domain adaptation},
  author={Liang, Jian and Hu, Dapeng and Feng, Jiashi},
  booktitle={International conference on machine learning},
  pages={6028--6039},
  year={2020},
  organization={PMLR}
}

\end{document}